\documentclass[letterpaper, 10 pt, conference]{ieeeconf}  

\IEEEoverridecommandlockouts                              

\usepackage{graphicx,subfigure}
\usepackage[linesnumbered,ruled,vlined]{algorithm2e}
\usepackage{microtype}

\usepackage{cite}  
\makeatletter
\def\@citex[#1]#2{\leavevmode
\let\@citea\@empty
\@cite{\@for\@citeb:=#2\do
{\@citea\def\@citea{,\penalty\@m\ }%
\edef\@citeb{\expandafter\@firstofone\@citeb\@empty}%
\if@filesw\immediate\write\@auxout{\string\citation{\@citeb}}\fi
\@ifundefined{b@\@citeb}{\hbox{\reset@font\bfseries ?}%
\G@refundefinedtrue
\@latex@warning
{Citation `\@citeb' on page \thepage \space undefined}}%
{\@cite@ofmt{\csname b@\@citeb\endcsname}}}}{#1}}
\makeatother

\title{\LARGE \bf
Multimodal Ensemble Approach to Incorporate Various Types of Clinical Notes for Predicting Readmission}

\author{Bonggun Shin,  Julien Hogan, Andrew B. Adams, Raymond J. Lynch, Rachel E. Patzer, and Jinho D. Choi$^{\dagger}$
	\thanks{B. Shin is with Department of Computer science, Emory University, Atlanta, GA 30303, USA.
bonggun.shin@emory.edu}%
	\thanks{J. Hogan is with Department of Surgery, Emory University, Atlanta, GA 30303, USA.
julien.hogan@emory.edu}%
	\thanks{A. B. Adams is with Department of Surgery, Emory University, Atlanta, GA 30303, USA.
andrew.b.adams@emory.edu}%
	\thanks{R. J. Lynch is with Department of Surgery, Emory University, Atlanta, GA 30303, USA.
ray.lynch@emoryhealthcare.org}%
	\thanks{R. E. Patzer is with Department of Surgery, Emory University, Atlanta, GA 30303, USA.
rpatzer@emory.edu}%
    \thanks{J. D. Choi is with Department of Computer science, Emory University, Atlanta, GA 30303, USA.
jinho.choi@emory.edu}%
	\thanks{$^{\dagger}$To whom correspondence should be addressed}%
}
	
\begin{document}

\maketitle
\thispagestyle{empty}
\pagestyle{empty}

\begin{abstract}

Electronic Health Records (EHRs) have been heavily used to predict various downstream clinical tasks such as readmission or mortality.
One of the modalities in EHRs, clinical notes, has not been fully explored for these tasks due to its unstructured and inexplicable nature.
Although recent advances in deep learning (DL) enables models to extract interpretable features from unstructured data, they often require a large amount of training data. 
However, many tasks in medical domains inherently consist of small sample data with lengthy documents; for a kidney transplant as an example, data from only a few thousand of patients are available and each patient's document consists of a couple of millions of words in major hospitals. 
Thus, complex DL methods cannot be applied to these kind of domains.
In this paper, we present a comprehensive ensemble model using vector space modeling and topic modeling.
Our proposed model is evaluated on the readmission task of kidney transplant patients, 
and improves 0.0211 in terms of c-statistics from the previous state-of-the-art approach using structured data, while typical DL methods fails to beat this approach.
The proposed architecture provides the interpretable score for each feature from both modalities, structured and unstructured data, which is shown to be meaningful through a physician's evaluation.

\end{abstract}

\section{INTRODUCTION}
\label{sec:intro}

\noindent Predicting post-discharge rehospitalization is one of the major research areas in health-informatics,
because the improvement of the prediction model could lead to better patient outcomes 
and efficient usages of medical resources~\cite{jencks2009rehospitalizations, harhay2013early}. 
According to Jones et al.~\cite{jones2016transitional}, 
about a half of surgical readmissions may be preventable,
indicating potential positive effects of the prediction model on both patients and medical institutes.
There have been many approaches to predict a readmission using electronic health records (EHRs), 
the majority of which are based only on structured data, 
such as demographic information, lab test values, 
and vital signs \cite{levy2001readmission,covert2016predicting, leal2017early}.

McAdams-Demarco et al.~\cite{mcadams2012early} found that 
a readmission of post-transplantation is a complex event consisting of  
various causes such as infections, rejection, 
and exacerbation of comorbidities.
For this reason, multi-modal features are more desirable when designing a prediction model,
implying that there is likely room for improvement in previous models by incorporating these rich untapped data.
Prior research has attempted to derive patterns from unstructured clinical notes,
as the field of natural language processing (NLP) is advancing.
These attempts enriched the model by extracting valuable information from unstructured data 
in predicting clinical outcomes~\cite{staff2013can}, 
or identifying patient phenotype cohorts~\cite{shivade2013review, zhou2014mining, shin2017classification}. 
Moreover, recent successes in deep learning (DL) have encouraged use of deep neural networks in
clinical NLP problems, and many of them have shown its superiority in various downstream clinical tasks
~\cite{shin2017classification,duggal2016predictive,craig2017predicting}.
However, these DL based models cannot be directly applied to many practical clinical problems
due to peculiarities of these clinical datasets:

\begin{itemize}
	\item Small sized samples - Many clinical downstream tasks consist of less than a few thousand data samples, which makes a model prone to be overfitted.
	\item Missing note types - Since not all types of notes are available for each patient, we need to impute the missing note modality, engendering an inevitable performance loss.
	\item Target-irrelevant sentences - Patients have multiple lengthy documents, consisting of a large number of words, however, only small portion of which are relevant to the target task. Therefore, important information tends to be diluted due to many non-informative words (or sentences). 
\end{itemize}

\noindent To overcome these issues, we propose an ensemble framework that doesn't require imputation of missing modalities, consisting of simple classifiers to circumvent overfitting, 
using vector space modeling and topic modeling as a feature to make it robust to long documents.

Our framework is evaluated on the Emory Kidney Transplant Dataset (EKTD), 
which task is to predict post-discharge rehospitalization at 30 days, 
being associated with poor outcomes of a patient.
The dataset comprises 80 structured variables along with three different types of clinical notes.
Our experiments show that 
the proposed framework outperforms the previous state-of-the-art approaches by 0.0211 in terms of c-statistics.
Not only that, our research further adds interpretability to the data by effectively incorporating
discriminating indices~\cite{shin2017wx} to the trained model.
To the best of our knowledge, this is the first time that 
an interpretable ensemble model is introduced for a readmission prediction problem with multiple modalities of input data.

\vspace{-1em}
\section{Approaches}
\subsection{Datasets}

\begin{table}[ht]
\centering
\begin{tabular}{c||r|r|r}
\bf Modality & \bf Patients & \bf Notes & \bf Common Patients \\
\hline \hline 
Structured                     & 2,060           & N.A.       & 2,060 \\
\hline
Consultations                  & 2,282           &  21,854    & 1,354                 \\
Progress                       & 2,444           & 202,296    & 1,415                 \\
Selection Conf. Ref.           & 2,843           &   3,512    & 2,033                 \\
\end{tabular}
\caption{Statistics of Structured and Unstructured dataset.}
\vspace{-3em}
\label{tbl:dataset}
\end{table}

\noindent The proposed framework is evaluated on the Emory Kidney Transplant Dataset (EKTD), after Institutional Review Board approval. 
It consists of 2,060 patients of structured data (80 predictors), and various number of patients depending on the type of clinical notes as summarized in Table~\ref{tbl:dataset}.
We utilized three common clinical notes:
Consultations including all notes redacting for outpatient consultations during the year prior to transplantation, 
Progress including all notes written during the transplant admission 
and Selection Conference summarizing the result of the pre-transplant screening and justifying the waiting-list registration.
As shown by the Table~\ref{tbl:dataset}, some patients don't have all types of notes,
indicating the need of imputation, if a typical ensemble classifier is used.
The target values, patient 30-days readmission outcomes are also recorded as a binary variable.
Of the final population, 633 (30.7\%) were rehospitalized after 30 days.

\subsection{Baseline Model with Structured Dataset}

\noindent Structured data include demographic and clinical characteristics of both the recipient and the donor, features related to labs results during the transplant admission.
Many lab test values are time series, but only the last value at the discharge is used to form a fixed length feature. 
All non-binary categorical variables are transformed into dummy-binary variables, 
increasing the feature length into 92. 
Standardized normalization are applied to each feature using the mean and standard deviation calculated from the training samples.
The baseline model is trained using only this structured features.

\subsection{Feature Representation for Unstructured Dataset}

\noindent EKTD contains three types of clinical notes. 
Notes written after the discharge are excluded for creating a fair model.
These notes are preprocessed using the ELIT tokenizer~\footnote{https://github.com/elitcloud/elit}, 
and all non-alphabetic tokens and typical English stopwords are removed from the notes.
If a patient has multiple notes in a specific note type, those notes are merged into one document to transform it to a fixed sized feature vector.
In this paper, vector space modeling and topic modeling are used as a vectorization method.

\subsubsection{Vector Space Modeling}

We use a popular vector space modeling, 
the term frequency-inverse document frequency (TF-IDF), 
since it can effectively filter out corpus specific stopwords, which are not covered by conventional English stopwords.
TF-IDF model is fitted using only training data for each fold, and the test data is vectorized using the fitted TF-IDF model.
The resulting TF-IDF vector has the size of $|V_n|$, where $V_n$ is the number of vocabulary of the note $n$.

\subsubsection{Topic Modeling}

Topic modeling is another way of document vectorization method. 
We use a topic distribution generated by 
Latent Dirichlet Allocation (LDA)~\cite{blei2003latent} to represent each type of notes of a patient.
As the successful clinical note processing work~\cite{ghassemi2014unfolding} suggested,
we used 50 topics. 
We set the hyperparameter, $\alpha=\frac{5}{numberTopic}$ after trying various values on each validation set.
A final topic distribution is drawn from a MCMC chain after trained 3,000 iterations.
The resulting LDA vector has the size of the number of topics, 50.

\subsection{Incorporating Unstructured Modalities}

\begin{figure}[htp!]
\centering     
\subfigure[Naive concatenation (Baseline).]
{\label{fig:lrconcat}\includegraphics[scale=0.65]{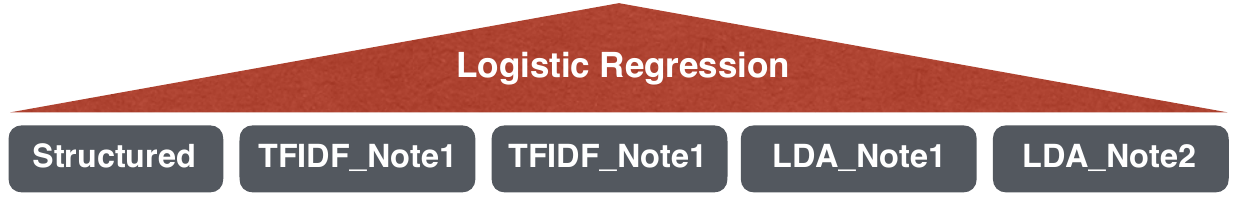}}
\subfigure[Averaged sigmoids (Proposed).]
{\label{fig:lravg}\includegraphics[scale=0.65]{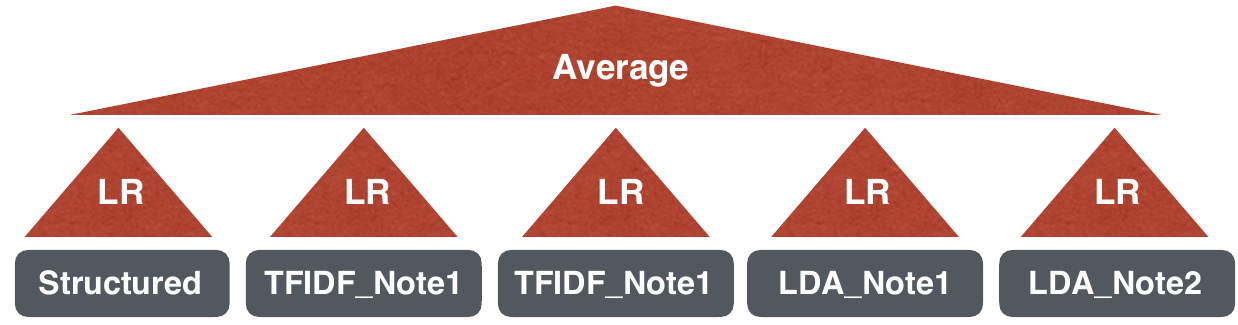}}
\caption{Multimodal ensemble models.}
\label{fig:ensemble}
\end{figure}

\noindent A naive way of incorporating unstructured modalities
is a logistic regression with 
naively concatenated features (Fig. ~\ref{fig:lrconcat}), 
where each feature comes from different modalities;
A structured feature ($x_s \in R^{92}$), TFIDF vectors ($x_{tfidf,n} \in R^{|V_{n}|}$), 
and LDA vectors ($x_{lda,n} \in R^{50}$), where $n$ represents a note type.
Note that this method requires imputation because all modalities should be concatenated together to form a fixed size vector.
However, the proposed multi-modal approach, 
the averaged sigmoids method (Fig. ~\ref{fig:lravg}), 
doesn't require imputation of the missing modalities, 
because each model for each modality is trained separately.
Predictions from these separate models are averaged into one final probability value.
Therefore, if some of the modalities are missing,
the proposed method just takes an average without those modalities.


\subsection{Feature Importance}
\noindent We apply the discriminative index (DI)~\cite{shin2017wx} to each modality 
to identify corresponding key features.
The DI algorithm simultaneously utilizes input features and associated coefficients
to calculate the contribution score, WX.
Shin et al.~\cite{shin2017wx} showed that
the scores calculated from both inputs and weights are more 
accurate than just considering weights.
For example, even if a weight value is big, it could not be one of the important features, because the reason why the weight became big is that
the corresponding input feature values might simply in a range of very small numbers.
By comparing the WX scores of two different cohorts (negative and positive samples),
proper feature impact scores can be calculated as described in Algorithm~\ref{alg:DIchi}.


\begin{algorithm}[hbtp!]{
	\SetAlgoLined
    \SetAlgoVlined
    \DontPrintSemicolon
	\KwIn{$X, Y, \theta, c$}
	\KwOut{Sorted Feature List}
	$\hat{X}^{True} = avg(X^{True})$;\\
	$\hat{X}^{False} = avg(X^{False})$;\\
	$WX^{True} = \theta^T \cdot \hat{X}^{True}$;\\
	$WX^{False} = \theta^T \cdot \hat{X}^{False}$;\\
	
	\For{$k \in \{1, \ldots, K\}$}
	{
	    $DI_{k} \leftarrow |WX^{True}_{k} - WX^{False}_{k}|$ 
	}

	\Return ${argsort} (DI)$
	\caption{\small DI~\cite{shin2017wx}}
	\label{alg:DIchi}
}
\end{algorithm}
\vspace{-1em}
	
	

\section{EXPERIMENTS}

\noindent We present the results of the proposed ensemble framework 
contrasting other frameworks
and show the effectiveness of integration of multimodal features for a readmission prediction task.
In addition, we provide the interpretability of the framework 
by showing the usefulness of the selected top-10 features of each modality.
5-fold cross validation was used to evaluate all approaches. 
For each fold, we held out a 20\% of the patients as a test set. 
The remaining 80\% of patients were used to vectorizations and training the models.

\subsection{Training Details}

\noindent The number of features of structured modality is 92, 
three topic modeling modalities are 50,
and three vector space modeling modalities vary depending on the training data of each subfold.
For instance, the first subfold finds the optimal vector dimensions as
30,275, 33,501, and 20,598 for 
Consultations, Progress, and Selection Conference, respectively.

\subsection{Performance Comparison}

\noindent The experiments are designed to validate that the proposed framework
successfully ameliorates three hurdles of practical clinical problems (Section~\ref{sec:intro}) including EKTD.
The first issue is overfitting.
As discussed by Shin et. al~\cite{shin2017classification},
unless the model is dealing with large dataset,
deep learning has no edge compared to a simple logistic regression model.
Reflecting this lesson, we pick a logistic regression as a classifier for all frameworks.
The second issue is possible performance degradation due to imputation of missing modalities.
We contrast two ensemble models, one that requires imputation (Fig~\ref{fig:lrconcat}),
and another one that doesn't (Fig~\ref{fig:lravg})
to show that how our framework (Fig~\ref{fig:lravg}) effectively handles missing modalities.
The last problem is information dilution caused by lengthy documents.
We compare the proposing vectorization method to 
the recent advanced vector modelings, word vectors~\cite{NIPS2013_5021} 
and document vectors~\cite{le2014distributed},
which shows the proposed one better capture the useful information in a low signal to noise environment.
For word vectors, we used the pre-trained biomedical word2vec~\cite{moen2013distributional},
and document vectors are trained on each training set of each note type.
We excluded convolutional~\cite{shin2017classification} and 
recurrent neural network~\cite{grnarova2016neural} based models, because the size of the model exceeds the memory capacity due to the huge number of tokens per one patients, ranging from one to ten millions.




\noindent Table~\ref{tbl:experiment} shows that the proposed ensemble framework, 
consisting of vectorspace modeling, topic modeling and averaged sigmoid ensemble classifier (Fig~\ref{fig:lravg}),
outperforms the other frameworks.
This framework successfully integrates three types of 
clinical notes with structured dataset,
improving 0.0211 compared to the structured only model.
Moreover, 95\% confidence interval of the proposed framework
indicates that our framework consistently outperforms the baseline by more than 0.01 margin.

\begin{table}[htbp!]
\begin{tabular}{l||ccc}
\multicolumn{1}{c||}{\bf Method} & \bf Avg. c-stats & \bf 95\% CI & \bf Delta \\ \hline \hline
Structured Only      & 0.6523  & (0.6218, 0.6829) & -            \\ \hline
Avg.W2V (Concat)     & 0.6561  & (0.631, 0.682)   & 0.0038        \\
Avg.W2V (Avg.Sig.)   & 0.6597  & (0.635, 0.684)   & 0.0074        \\
Doc2Vec (Concat)     & 0.6522  & (0.624, 0.681)   & -0.0001        \\
Doc2Vec (Avg.Sig.)   & 0.6491  & (0.624, 0.674)   & -0.0032        \\
TFIDF-LDA (Concat)   & 0.6669  & (0.6488, 0.6850) & 0.0146        \\
TFIDF-LDA (Avg.Sig.) & \textbf{0.6734}  & (0.6635, 0.6834) & \textbf{0.0211}
\end{tabular}
\caption{Averages and 95\% conf. intervals of c-stats of all five folds. 
Top most row represents the scores of the baseline with only structured features.
Others are combinations of two multimodal ensembles with different vectorizations.}
\label{tbl:experiment}
\end{table}

\subsection{Feature Analysis}
\begin{figure*}[hbpt!]
\centering     
\subfigure[Structured]
{\label{fig:fa1}\includegraphics[height=70pt]{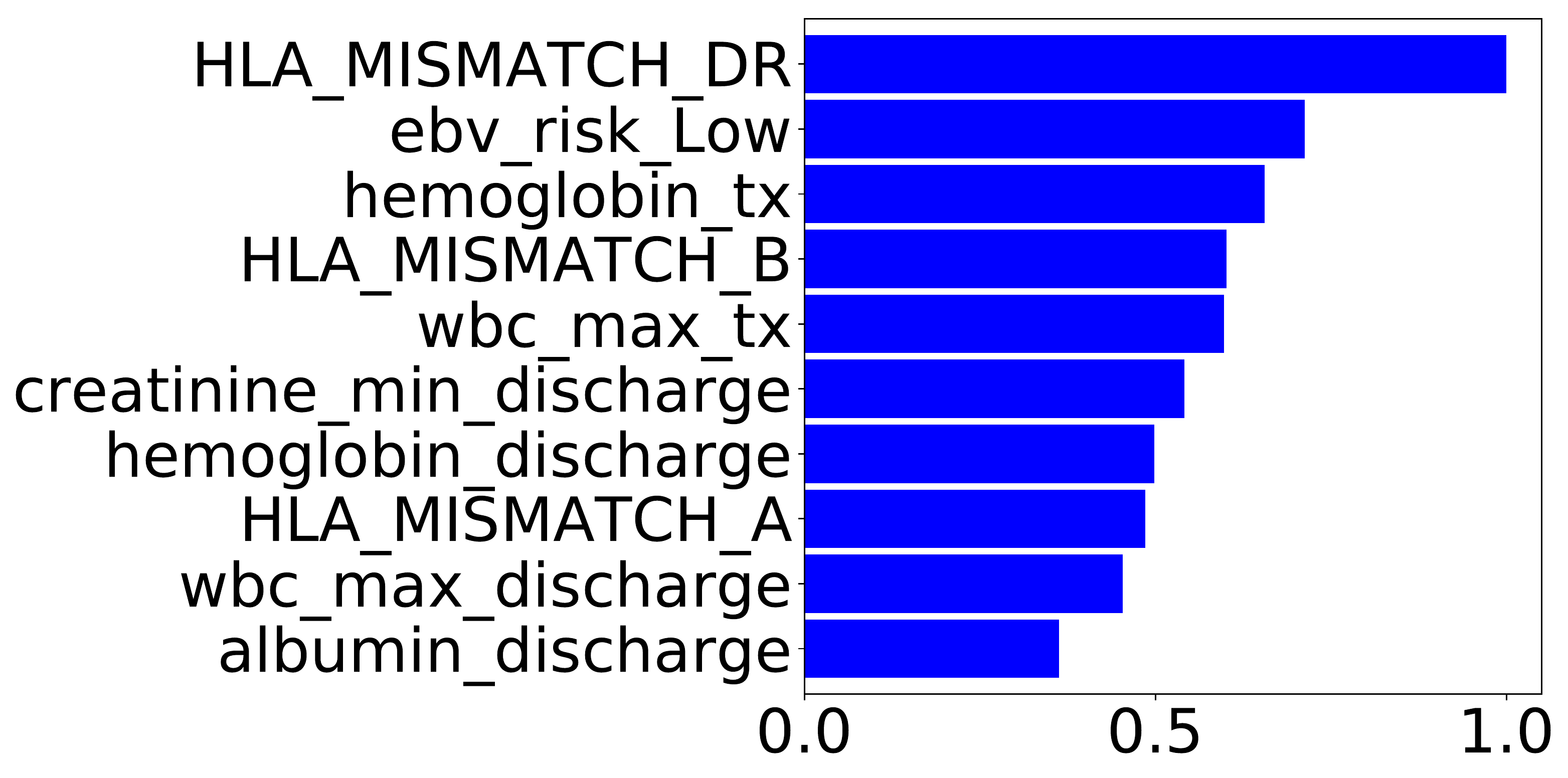}} 
\subfigure[TFIDF Consultations]
{\label{fig:fa2}\includegraphics[height=70pt]{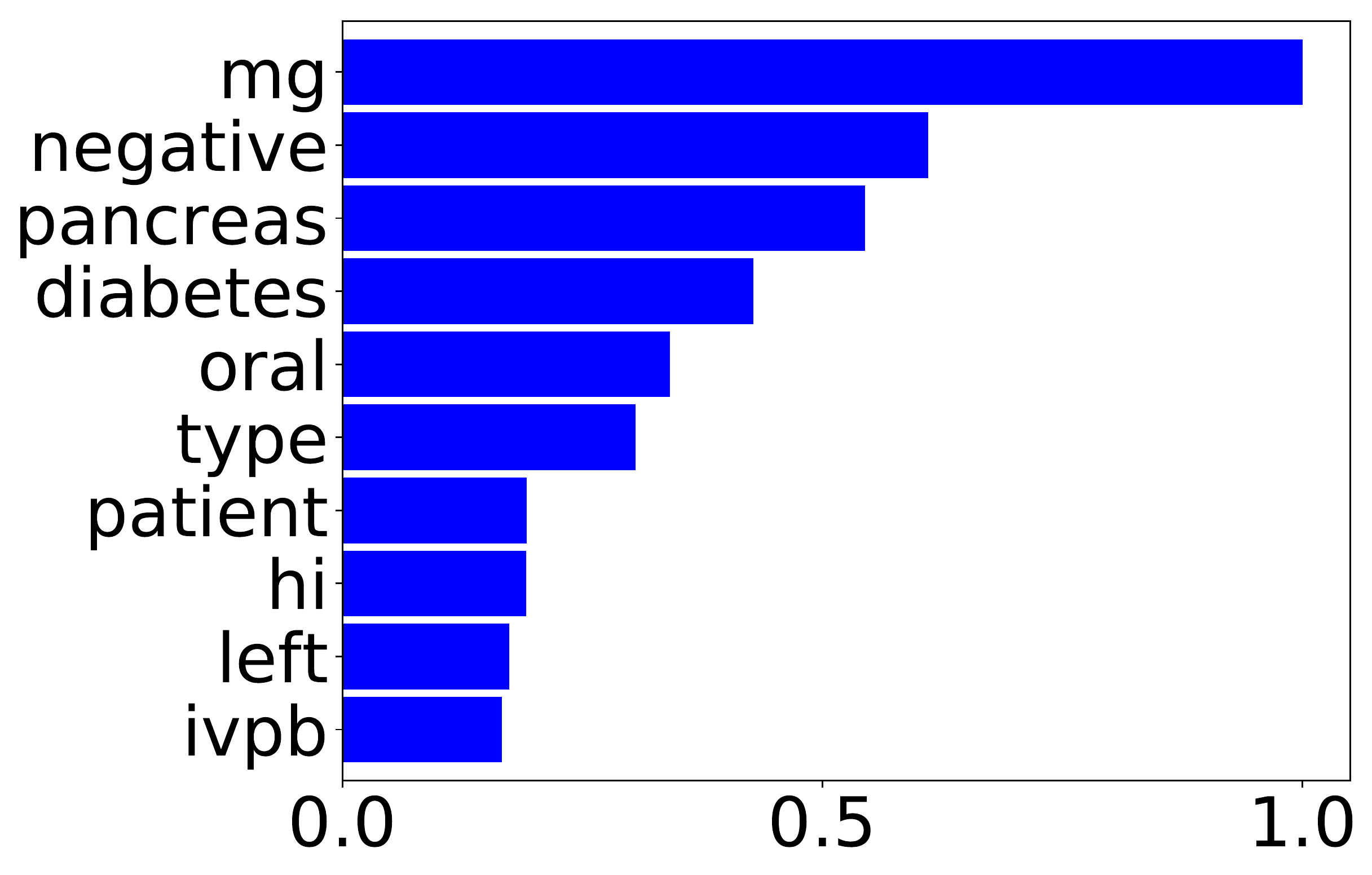}} 
\subfigure[TFIDF Progress]
{\label{fig:fa3}\includegraphics[height=70pt]{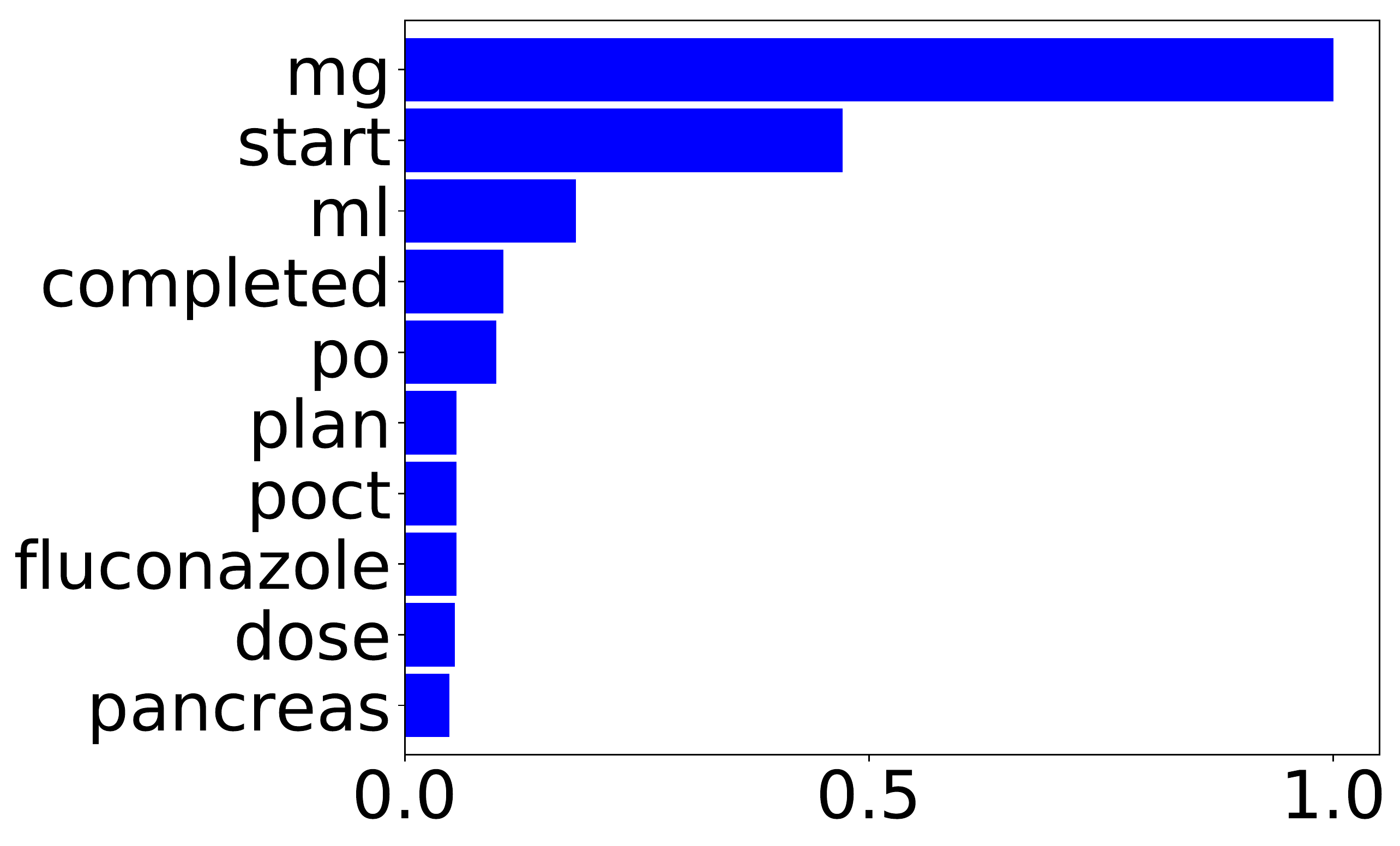}} 
\subfigure[TFIDF Selection Conference]
{\label{fig:fa4}\includegraphics[height=70pt]{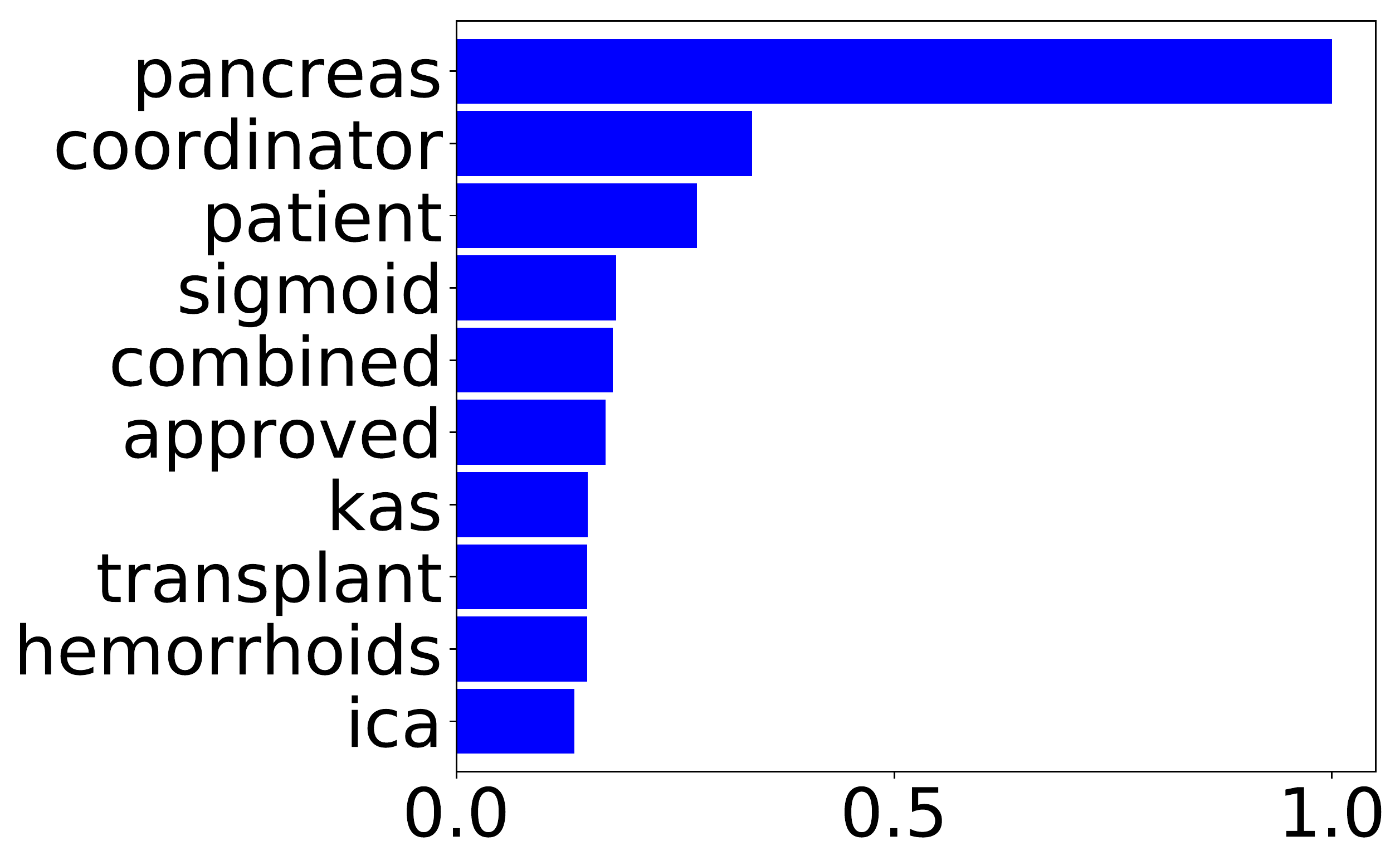}} 
\subfigure[LDA Consultations] 
{\label{fig:fa5}\includegraphics[width=150pt]{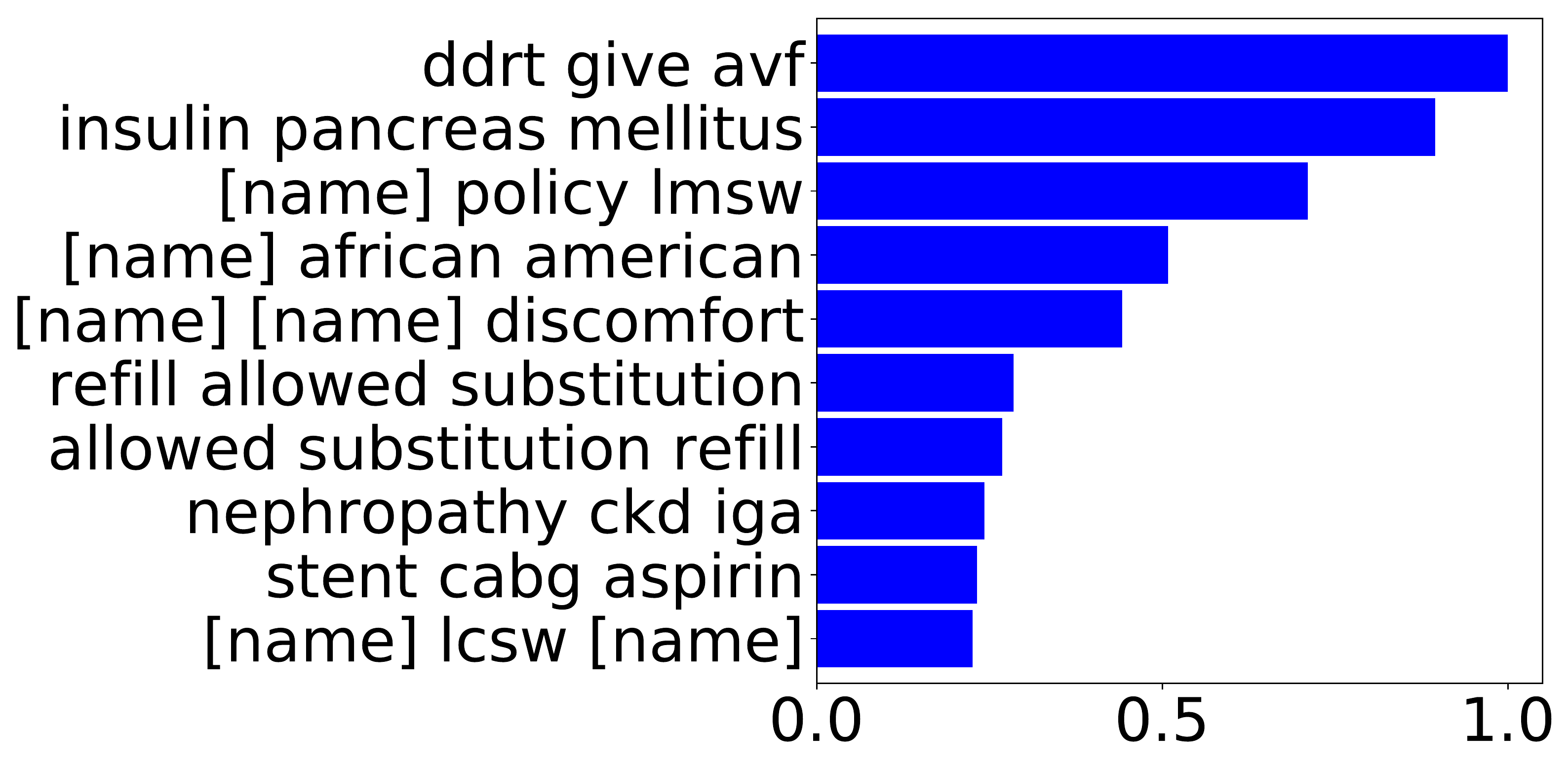}} 
\subfigure[LDA Progress]
{\label{fig:fa6}\includegraphics[width=150pt]{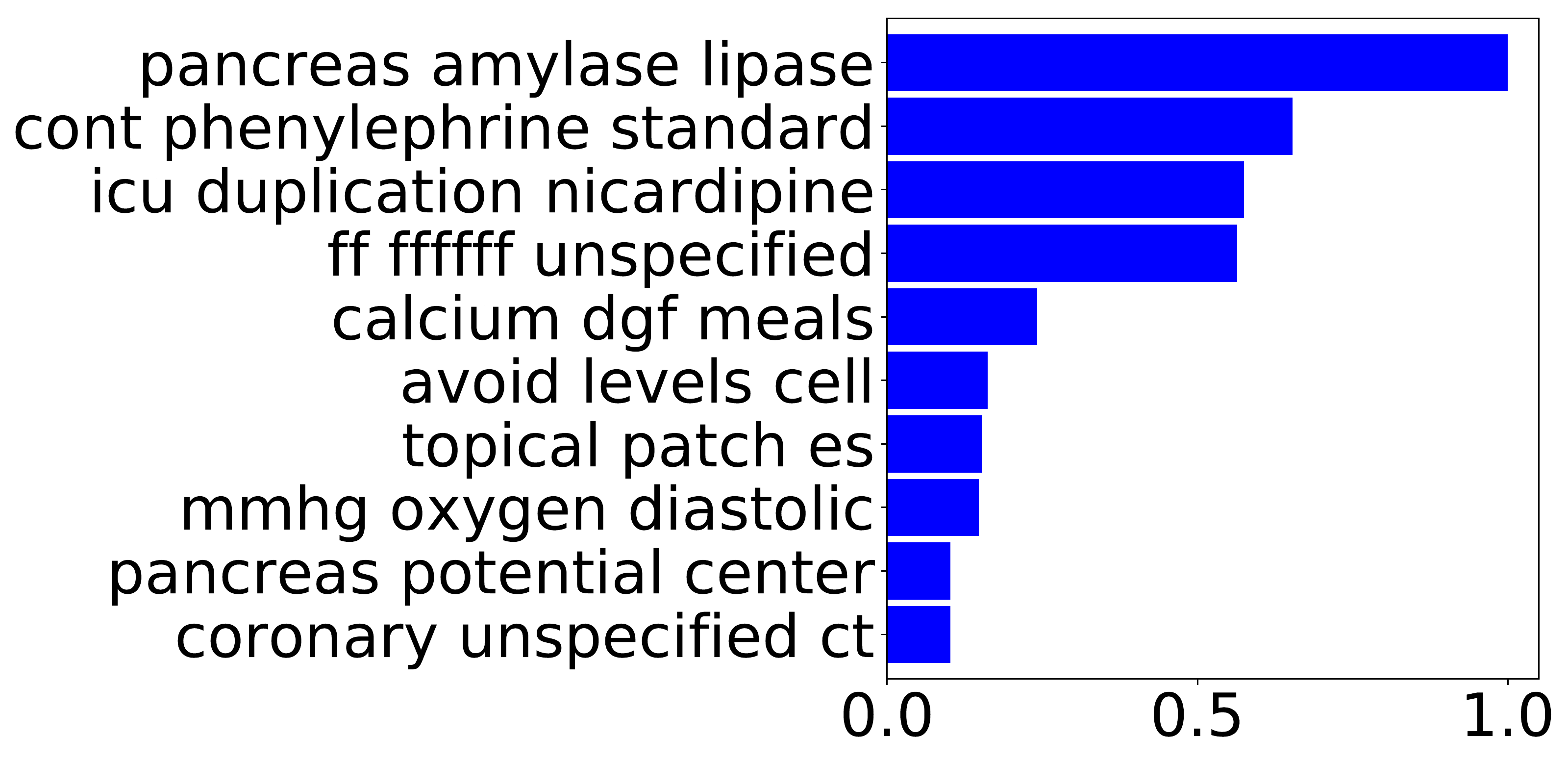}} 
\subfigure[LDA Selection Conference]
{\label{fig:fa7}\includegraphics[width=150pt]{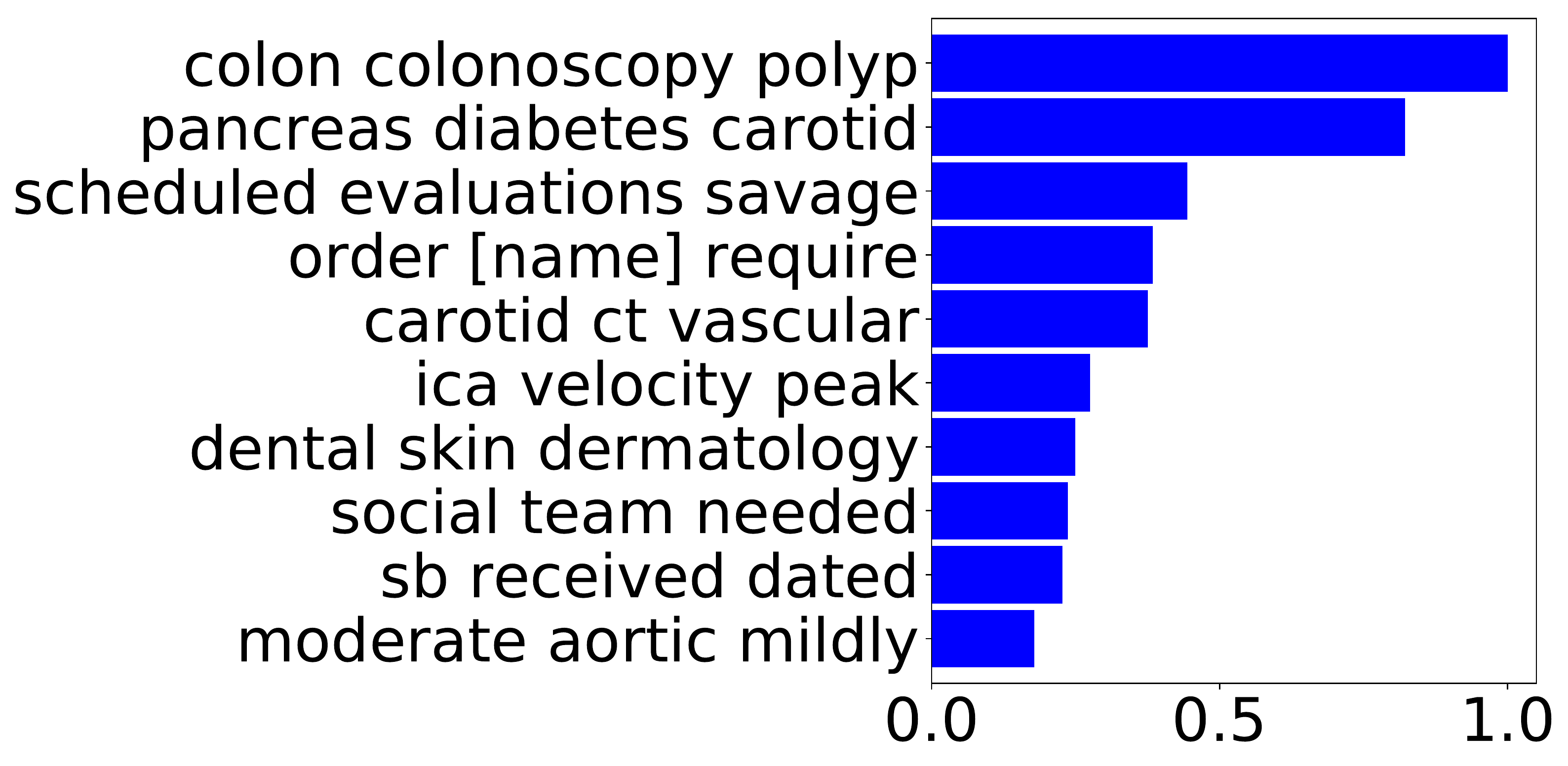}} 
\caption{Top 10 important features, which are min-max normailzed. For LDA modalities, top three frequent words for each topic are listed on the y-axis. All names of a physician or a social worker are replced to ``[name]''.}
\label{fig:di}
\vspace{-1.5em}
\end{figure*}

\noindent Algorithm~\ref{alg:DIchi} is applied to all seven modalities, 
and Top-10 important features are presented in Figure~\ref{fig:di}.
Top predictors from the structured data included mostly labs results, 
such as hemoglobin, albumin, and creatinine level which is a marker of the function of the transplanted kidney. The overall predictive accuracy of this model was low and consistent with previously published predictive models of 30-day readmission. Other important predictors are the quality of the immunological matching between the donor and the recipient (HLA\_MISMATCH). 
The 6 models based on clinical notes captured relevant predictors of hospital readmission,
including assorted patients’ comorbidities.
All models' selected terms are related to diabetes or diabetes-related complications 
(pancreas, diabetes, insulin, and mellitus). 
Similarly, all LDA models reported topics related to cardio-vascular complications (carotid, coronary stent, and coronary artery bypass grafting) 
and digestive neoplasia (colonoscopy, polyp, and sigmoid). 

Severity markers were also extracted as major predictors within the progress notes indicating the need for admission in the intensive care unit (ICU), respiratory failure (oxygen) or the need for intravenous medications to either lower (nicardipine) or increase (phenylephrine) blood pressure (mmHg). Of specific interest were the topics related to socio-economic status or social support. Indeed, if patients’ comorbidities are usually captured in classic structured databases, social-economic features are often poorly recorded. The need for a social evaluation of the patients expressed either in the selection conference notes (“social team needed”) or in the consultations as demonstrated by the presence of social workers names within the notes (Licensed Master Social Worker (LMSW), Licensed Clinical Social Worker (LCSW)) was included as a top predictive feature in 2 out of 3 LDA-based models. Similarly, topics related to medication delivery and adherence were also captured (“refill allowed substitution”). This finding is extremely interesting since much emphasis is currently made on the impact of adherence on clinical outcome and underline the potential of NLP in this field.

\section{CONCLUSIONS}

\noindent This paper proposes a multi-modal ensemble framework with
vector space and topic modeling features 
that effectively integrates structured and unstructured dataset for predicting
readmissions at 30 days.
Our experiments show that
this framework not only adequately handles missing modality,
but properly catches useful information from a very long document.
In addition, we introduced a way to interpret the prediction results, 
which could potentially be valuable 
in medical actions.
The physician’s evaluation showed that the provided framework is meaningful 
in that it not only allows the extraction of both classic predictors previously reported in other studies, 
but also the predictive features covers fields that are usually poorly covered in structured databases such as socio-economic status, social support or medication delivery and adherence.

 \vspace{-0.5em}





\bibliographystyle{IEEEtran}
\bibliography{IEEEabrv,bhi_ensemble}

\end{document}